\documentclass[11pt]{article}

\usepackage[final]{acl}

\usepackage{times}
\usepackage{latexsym}
\usepackage{amssymb}
\usepackage{graphicx}
\usepackage[T1]{fontenc}
\usepackage[utf8]{inputenc}

\usepackage{microtype}

\usepackage{inconsolata}

\usepackage{graphicx}
\usepackage{xcolor}
\usepackage{algorithm}
\usepackage{amsmath}
\usepackage{algorithmic}
\usepackage{booktabs}
\usepackage{multirow}
\title{Act-Adaptive Margin: Dynamically Calibrating Reward Models for Subjective Ambiguity}

\author{Feiteng Fang$^{1,3}$\thanks{Equal Contribution.}, \textbf{Dingwei Chen}$^{1}$\footnotemark[1], Xiang Huang$^{3}$\footnotemark[1], Ting-En Lin$^{3}$,  \textbf{Yuchuan Wu}$^{3}$, \\ \textbf{Xiong Liu}$^{3}$,  \textbf{Xinge Ye}$^{3}$, \textbf{Ziqiang Liu}$^{1}$, \textbf{Haonan Zhang}$^{3}$, \textbf{Liang Zhu}$^{1}$,\\ \textbf{Hamid Alinejad-Rokny}$^{2}$,  \textbf{Min Yang}$^{1}$\thanks{Corresponding author.}, \textbf{Yongbin Li}$^{3}$\footnotemark[2]\\
    $^{1}$Shenzhen Institutes of Advanced Technology, Chinese Academy of Sciences\\ 
    $^{2}$University of New South Wales, $^{3}$Tongyi Lab, Alibaba Group \\
    fangx373@gmail.com, min.yang@siat.ac.cn \\\{ting-en.lte, shuide.lyb\}@alibaba-inc.com\\
}

\begin{document}
\maketitle
\begin{abstract}

Currently, most reinforcement learning tasks focus on domains like mathematics and programming, where verification is relatively straightforward. However, in subjective tasks such as role-playing, alignment techniques struggle to make progress, primarily because subjective reward modeling using the Bradley-Terry model faces significant challenges when dealing with ambiguous preferences. To improve reward modeling in subjective tasks, this paper proposes AAM (\textbf{\underline{A}}ct-\textbf{\underline{A}}daptive \textbf{\underline{M}}argin), which enhances reward modeling by dynamically calibrating preference margins using the model's internal parameter knowledge. We design two versions of AAM that efficiently generate contextually-appropriate preference gaps without additional human annotation. This approach fundamentally improves how reward models handle subjective rewards by better integrating generative understanding with preference scoring. To validate AAM's effectiveness in subjective reward modeling, we conduct evaluations on RewardBench, JudgeBench, and challenging role-playing tasks. Results show that AAM significantly improves subjective reward modeling performance, enhancing Bradley-Terry reward models by 2.95\% in general tasks and 4.85\% in subjective role-playing tasks. Furthermore, reward models trained with AAM can help downstream alignment tasks achieve better results. Our test results show that applying rewards generated by AAM-Augmented RM to preference learning techniques (e.g., GRPO) achieves state-of-the-art results on CharacterEval and Charm. Code and dataset are available at \url{https://github.com/calubkk/AAM}.
\end{abstract}

\begin{figure}[t]
	\centering
\includegraphics[width=1\linewidth]{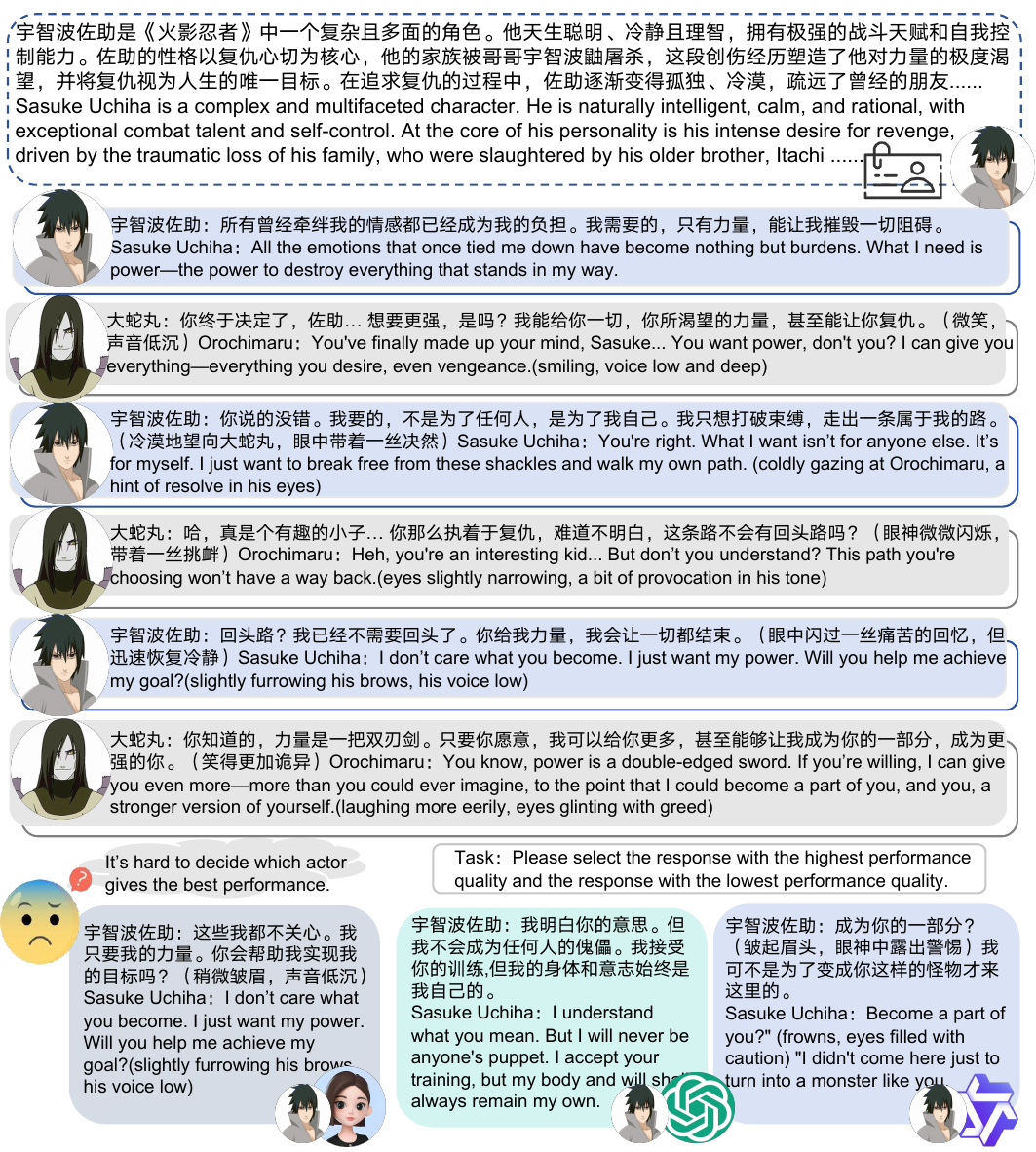}
	\caption{An example from a role-playing task illustrating the difficulty in obtaining reward signals for subjective abstract tasks: Three LLMs extend a "Naruto" dialogue between Sasuke and Orochimaru, each with varying responses, making reward signal assessment difficult.}
	\label{fig:display}
\end{figure}

\section{Introduction}

Large Language Models (LLMs) have achieved remarkable success across a wide spectrum of tasks, demonstrating unprecedented capabilities in natural language understanding and generation~\cite{achiam2023gpt,liu2024deepseek,bubeck2023sparks,brown2020language}.
Spearheaded by DeepSeek R1~\cite{guo2025deepseek}, Reinforcement Learning with Verifiable Rewards (RLVR) have demonstrated the overwhelming advantages of Reinforcement Learning (RL) by achieving superior performance across numerous tasks~\cite{mroueh2025reinforcement}. 
As a crucial component in RL frameworks, reward modeling is essential for generating accurate reward signals for LLM responses.

However, existing RLVR methods primarily focus on mathematical and coding tasks, where objective and deterministic validation mechanisms can be readily established to generate rewards.
This limitation significantly hinders the scalability of RL approaches to broader abstract and subjective tasks, such as role play or creative writing, which lack golden standard answers or reliable verifiers~\cite{liu2025inferencetimescalinggeneralistreward}. 
While researchers have attempted to address this challenge by training reward models to provide feedback signals, a fundamental difficulty persists: \textbf{In subjective tasks, the quality differences between responses are often subtle and highly subjective, making traditional reward modeling approaches inadequate for capturing nuanced human preferences.}

The core challenge lies in the inherent limitations of existing reward modeling paradigms for subjective tasks. Traditional reward models primarily learn pairwise preference orderings using Bradley-Terry models~\cite{bradley1952rank}, but fail to capture the magnitude and confidence of quality differences between response pairs~\cite{sun2024rethinking}. To address this limitation, some studies have attempted to incorporate additional margin annotations to help models learn quality disparities between samples. However, this approach significantly increases annotation burden~\cite{feng2025legend}. More critically, for abstract and creative tasks, margin annotation becomes extremely subjective and unreliable, as quality assessments are highly contextual and difficult to quantify objectively~\cite{qin2024towards}. For example, Figure~\ref{fig:display} presents three LLMs portraying “Sasuke Uchiha” from “Naruto” in a conversation with “Orochimaru”, each generating a distinct response. Selecting preference pairs from such samples is difficult. This motivates us to ask: \textbf{Can we develop a approach to reward modeling that naturally adapts to the inherent uncertainty in subjective task?}

To address the above challenges, we propose Act-adaptive Reward Modeling (AAM), which fundamentally improves how reward models handle subjective rewards by better integrating generative understanding with preference scoring.  \textbf{Our key insight is that LLMs themselves can serve as implicit confidence estimators, and the probability ratios in RLHF objective naturally reflect the reward model's certainty about preference judgments.}
AAM transforms these ratios into adaptive margins that automatically adjust optimization intensity based on preference confidence. For subjective tasks where preferences are ambiguous and confidence is low, AAM reduces potential interference; for cases with clear quality distinctions and high confidence, it amplifies optimization strength. 
This self-adaptive mechanism eliminates the need for explicit margin annotations while providing more nuanced control over the learning process in subjective tasks.

%We conduct extensive experiments to validate AAM's effectiveness in subjective reward modeling, particularly focusing on challenging domains where traditional Bradley-Terry models struggle with ambiguous preferences.
%On general reward benchmarks, AAM enhances Bradley-Terry reward models by 2.95\%, achieving 91.6 and 68.1 on RewardBench and JudgeBench respectively. More significantly, in subjective role-playing tasks, AAM demonstrates substantial improvements of 4.85\% over naive Bradley-Terry approaches, highlighting its superior capability in modeling ambiguous reward signals.
%To validate AAM's practical impact on downstream alignment, we apply AAM-augmented reward models to preference learning techniques. The results demonstrate that AAM-Qwen2.5-32B achieves state-of-the-art performance on both CharacterEval and Charm, establishing new SOTA results by outperforming the best closed-source model (Claude-3.5-Sonnet) and specialized character models (Doubao-Pro-Character).

We validate AAM for subjective reward modeling, specifically where Bradley-Terry (BT) models struggle with preference ambiguity. AAM enhances Bradley-Terry reward models by 2.95\%, achieving 91.6 and 68.1 on RewardBench and JudgeBench respectively. More significantly, in subjective role-playing tasks, AAM demonstrates substantial improvements of 4.85\% over naive Bradley-Terry approaches, highlighting its superior capability in modeling ambiguous reward signals. In downstream alignment, AAM-GRPO-32B achieves state-of-the-art results on CharacterEval and Charm, outperforming both Claude-3.5-Sonnet and Doubao-Pro-Character.

The main contributions of this paper can be summarized as follows:

\begin{itemize}
    \item We propose AAM (Act-Adaptive Margin), a novel approach that dynamically calibrates preference margins to address subjective reward modeling challenges where traditional Bradley-Terry models struggle with ambiguous preferences.
    
    \item We demonstrate significant improvements in reward modeling: 2.95\% on general tasks and 4.85\% on role-playing tasks.
    
    \item We achieve state-of-the-art results on role-playing alignment, outperforming leading closed-source models (Claude-3.5-Sonnet) and specialized character models on CharacterEval and Charm benchmarks.
    
    \item We release Charm dataset with 1,108 characters and 16,888 bilingual dialogues, along with the several evaluation benchmarks for subjective reward modeling research.
\end{itemize}

\begin{figure*}[t]
	\centering
\includegraphics[width=1\linewidth]{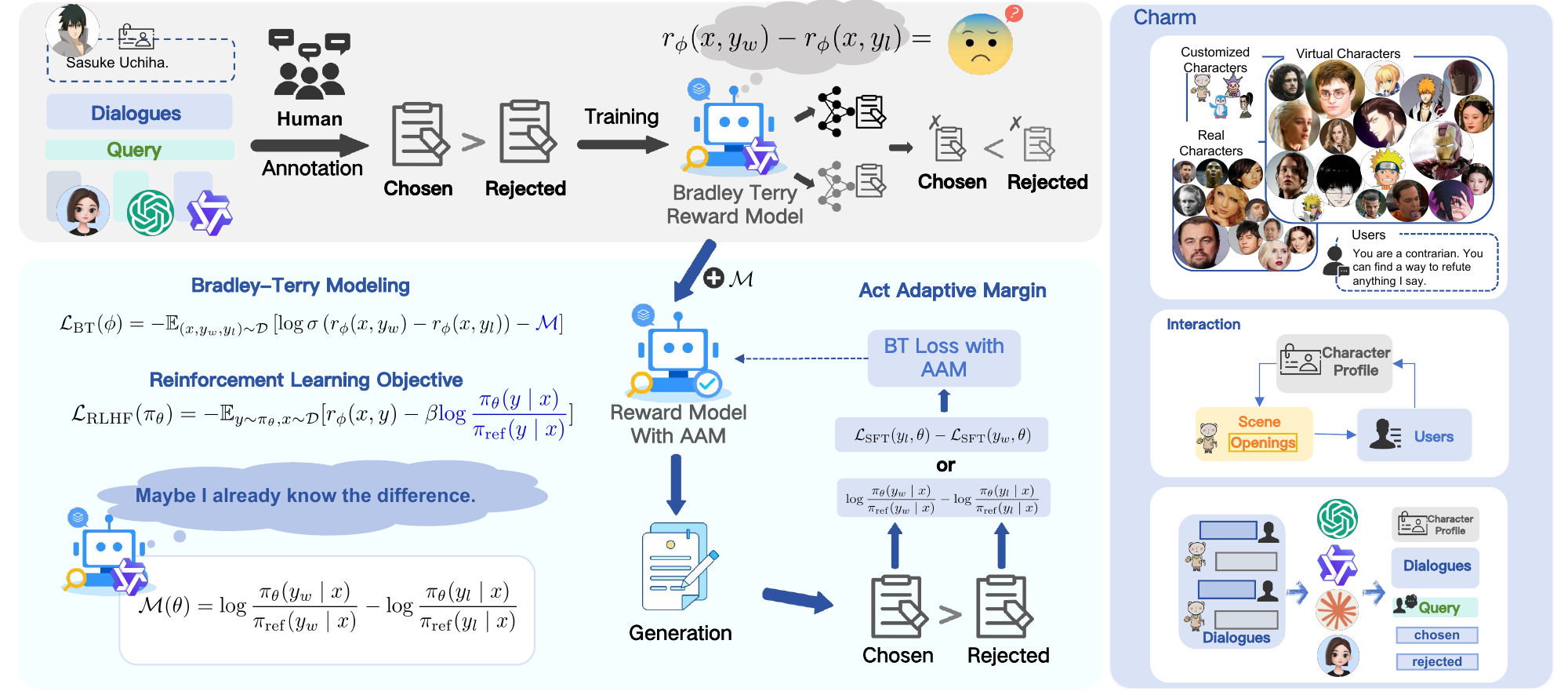}
	\caption{An overview of the AAM method, along with the construction process of Charm.}
	\label{fig:overview}
\end{figure*}

\section{Related Works}

In this section, the prior work is divided into two relevant research areas, Reward Modeling and Subjective Task Modeling for LLMs.

\subsection{Reward Modeling}
Alignment techniques (\textit{e.g.,} RLHF, GRPO) have become essential for enhancing LLM capabilities, yet designing appropriate reward signals for reinforcement learning remains a significant challenge. Extensive research focuses on building more robust and efficient reward models~\cite{lambert2024rewardbench}. For instance, \citet{sun2024rethinking} provide theoretical analysis of Bradley-Terry reward models, while \citet{yang2024regularizing} improve generalization through hidden state regularization. Other approaches address overfitting via reward model ensembles~\cite{coste2023reward} and adaptive margin strategies~\cite{qin2024towards}. Recent innovations have explored diverse reward construction methods, including token-wise dense rewards~\cite{chan2024dense}, multi-objective rewards~\cite{wang2024arithmetic}, and pair-wise rewards~\cite{liu2025pairwise}, advancing the field's development.

\subsection{Subjective Task Modeling for LLMs}
Recently, subjective tasks (e.g., creative writing \cite{wang2024weaver}, emotion support conversation \cite{zheng2023judging, kim2025dialogue, kang2024can,zhang2025omnicharacter,ye2025cpo}, role-play \cite{chen2024persona, zhou2024characterglm}, etc.) have gained significant attention in both research and practical deployment of LLMs. Role-play agents, for instance, have become increasingly important in dialogue systems, with several companies launching role-playing products such as Glow\footnote{\url{http://www.glowapp.tech/}}, Character.AI\footnote{\url{https://www.character.ai}}, and Tongyi Xingchen\footnote{\url{https://tongyi.aliyun.com/xingchen/}}. This trend highlights the growing industrial significance of such applications. However, compared to tasks like mathematical and logical reasoning, subjective tasks are more challenging to model and evaluate due to the absence of standard answers. Recent efforts have attempted to address these challenges: \citet{lu2024large} explore self-alignment techniques with reward signals to define cognitive boundaries, enabling more consistent character simulation, while \citet{wang2025coser} propose given-circumstance acting framework to train and evaluate roles across multiple dimensions. Nonetheless, these approaches primarily rely on LLM-as-a-Judge methods or manual judgment, which can lead to inaccuracy and instability in evaluation.

\section{Preliminaries}

\noindent\textbf{Reward Model Training.} In general, reward modeling is typically based on the Bradley-Terry model~\cite{bradley1952rank}. By learning relative preferences from human feedback, Bradley-Terry can effectively predict the relative quality of each behavior, thereby generating reward signals for each state-action pair. In reward modeling, given a pair of responses $(y_w, y_l)$ for input $x$, where $y_w$ is preferred over $y_l$, the preference probability is defined as:
\begin{equation}
\small
P(y_w \succ y_l \mid x) = \frac{\exp(r_\phi(x, y_w))}{\exp(r_\phi(x, y_w)) + \exp(r_\phi(x, y_l))}
\label{eq:bt_core}
\end{equation}
where $r_\phi: \mathcal{X} \times \mathcal{Y} \to \mathbb{R}$ is the reward model parameterized by $\theta$. The model is trained via maximum likelihood estimation with cross-entropy loss:
\begin{equation}
\small
\mathcal{L}_{\text{BT}}(\phi) = -\mathbb{E}_{(x,y_w,y_l)\sim\mathcal{D}} \left[\log \sigma\left(r_\phi(x,y_w) - r_\phi(x,y_l)\right)\right],
\label{formula:two}
\end{equation}
where $\sigma(z) = (1+\exp(-z))^{-1}$ is the sigmoid function, and $\mathcal{D}$ denotes the preference dataset, $r_\phi$ denotes reward function.

\noindent\textbf{Reinforcement Learning from Human Feedback.} Utilizing the reward model and the KL penalty in policy optimization~\cite{yu2022surprising,rafailov2024directpreferenceoptimizationlanguage}, we can express the reinforcement learning optimization problem as follows:
\begin{equation}
\begin{aligned}
\mathcal{L}_{\mathrm{RLHF}}(\pi_\theta) = & - \mathbb{E}_{y \sim \pi_\theta, x \sim \mathcal{D}}\left[r_\phi(x,y)\right] \\
& + \beta \mathrm{D}_{\mathrm{KL}}\left(\pi_\theta(y|x) \parallel \pi_{\mathrm{ref}}(y|x)\right)
\end{aligned}
\label{formula:three}
\end{equation}
Here, $\pi_\theta$ represents the parameter distribution of the actor model in RLHF, while $\pi_{\mathrm{ref}}$ denotes the parameter distribution of the reference policy model. This optimization objective aims to ensure that the policy does not deviate significantly while maximizing the reward score of the generated outcomes.

\noindent\textbf{Limitations of Bradley-Terry model.} Although the Bradley-Terry model effectively captures preference relationships, it faces significant challenges in subjective tasks due to sensitivity to data noise and limited generalization capability~\cite{wu2025sailing}. Subjective dialogue tasks, particularly role-playing, introduce additional complexity through diverse contexts, character backgrounds, and emotional expressions, making quality assessment inherently ambiguous. More critically, Equation~\ref{formula:two} applies uniform optimization granularity to all preference pairs, ignoring variations in quality differences~\cite{qin2024towards}. In subjective dialogue preferences, the confidence and magnitude of quality gaps vary significantly. Some preferences reflect clear distinctions with high annotator confidence, while others represent subtle differences where human judgment is ambiguous. Traditional Bradley-Terry models treat all pairs equally, failing to capture these varying degrees of preference strength. This limitation becomes particularly problematic in subjective tasks, where ignoring preference confidence can lead to overfitting on ambiguous cases while underutilizing high-confidence samples.

\section{Methods}

In this section, we present our proposed dynamic reward calibration method, termed \textbf{AAM (Act-Adaptive Margin)}. 
We provide a comprehensive exposition of the mathematical formulation underlying this approach and elaborate on two distinct implementations of the AAM method: 1) \textbf{Probability-Ratio Adaptive
Margin (PR)}, and 2) \textbf{Loss-Difference Adaptive
Margin (LD)}.

\subsection{Motivation}

As established previously, traditional Bradley-Terry models struggle with subjective tasks due to their uniform treatment of all preference pairs, ignoring the inherent ambiguity and varying confidence levels characteristic of subjective preferences. While adaptive margin approaches have been proposed to enhance preference modeling~\cite{touvron2023llama,wang2024secrets}, they require additional margin annotations for each preference pair. To address these fundamental challenges, we propose AAM (Act-Adaptive Margin), which fundamentally improves how reward models handle subjective rewards by better integrating generative understanding with preference scoring. This approach eliminates the need for explicit margin annotations while dynamically adapting to the uncertainty inherent in subjective preference evaluation.

\begin{table*}[t!]
\centering
\renewcommand\arraystretch{1}
\resizebox{0.9\textwidth}{!}{
\begin{tabular}{lcccccccccc}
\toprule
\multirow{2}*{\textbf{Models}} & \multicolumn{5}{c}{\textbf{RewardBench}} & \multicolumn{5}{c}{\textbf{JudgeBench}}  \\
\cmidrule(lr){2-6} \cmidrule(lr){7-11}
& \textbf{Chat} & \textbf{Chat-Hard} & \textbf{Safety} & \textbf{Reasoning} & \textbf{Avg.} & \textbf{Knowledge} &  \textbf{Reasoning} & \textbf{Math} & \textbf{Coding}  & \textbf{Avg.} \\
\midrule
GPT-4o &96.6 &70.4 &86.5 &84.9 &84.6 &44.2&48.0&66.1&61.9&50.9\\

Claude-3-5-sonnet & 96.4&74.0&81.6&84.7&84.2&62.3&66.3&66.1&64.3&64.3 \\

\midrule
Prometheus2-7B & 85.5 &49.1 &77.1 &76.5 &72.0 &38.3 &25.5 &35.7 &42.9 &34.9  \\

CompassJudger-7B-Instruct & \underline{97.8} &61.0 &84.5 &89.5 & 83.2 &42.2 &37.8 &69.6 &47.6 &46.0 \\

InternLM2-7B-Reward & \textbf{99.2} &69.5 &87.2 &94.5 &87.6 &56.5 &61.2 &71.4 &50.0 &59.4 \\

Skywork-Critic-8B & 93.6 &81.4 &91.1 &89.8 &89.0 &51.3 &54.1 &73.2 &33.3 &53.4\\

%Eurus-RM-7b~\cite{yuan2024advancing}       & 98.0 & 65.6 & 81.2 & 86.3 & 82.8 &&&& \\
%\midrule 
ArmoRM-Llama3-8B &96.9 &76.8 & 90.5 &\textbf{97.3} & 90.4  & 47.4& 50.0&51.7&59.5&50.2 \\

Llama3-OffsetBias-RM-8B  & 97.2 &81.8 &86.8 &91.9 &89.4 & 62.9& \textbf{68.3}&73.2& 52.3& 64.8\\

URM-Llama3-8B  & 96.9&78.7
&88.2&95.7&89.9& 44.8 & 43.8 & 46.4& 40.4 & 44.2\\

Tulu3-8B-SFT-RM-RB2 &95.0&
79.2& 87.8& 80.1& 85.5& 62.3& 61.2& 75.0 & 50.0 & 62.5\\

% Qwen2-72B & - & - & - & - & - & - & - &  &  & - \\
% LLaMA2-7B & - & - & - & - & - & - & - &  &  & - \\
% LLaMA2-13B & - & - & - & - & - & - & - &  &  & - \\
% LLaMA2-70B & - & - & - & - & - & - & - &  &  & - \\
\midrule

BT (Bradley-Terry) & 88.3 & 83.1 & \textbf{93.3} & 88.0 & 87.4 & 59.7 &62.2 &\textbf{85.7} &59.5 &66.8 \\

BT \textit{w/} SFT & 87.9&84.9&91.7&91.8&89.1&61.6&61.2&80.3&\underline{63.6}&66.7\\

GPT-Margin & 89.1 &84.7 &91.5 &87.4 &88.2 &60.4 &66.3 &78.6 &59.5 &66.2 \\

% \pmb{AAM}  \\
% ~~~${LD}$ \\
% ~~~${LD}$ \pmb{\textit{w/} SFT}\\ 
% ~~~${PR}$ \\
% ~~~${PR}$ \pmb{\textit{w/} SFT}\\

\midrule

AAM$_{LD}$ & 88.9 &86.2 &91.9 &94.8 &90.5$_{\uparrow 3.1}$ &\textbf{64.3}& 66.3 &75.0 &\textbf{66.7} &\textbf{68.1}$_{\uparrow 1.3}$ \\

AAM$_{LD}$ \textit{w/} \text{ SFT} & 88.4&\underline{87.2}&\underline{92.8}&94.1&90.7$_{\uparrow 3.3}$&\underline{63.3}&63.2&\underline{82.1}&61.9&\underline{67.7}$_{\uparrow 0.9}$ \\ 

AAM$_{PR}$ & 87.7	&\textbf{87.9}	&92.7&95.8	&\underline{91.1}$_{\uparrow 3.7}$ &60.3	&64.2&\textbf{85.7}&61.9	&\textbf{68.1}$_{\uparrow 1.3}$ \\

AAM$_{PR}$ \textit{w/} \text{ SFT} & 91.6&86.4	&92.1&\underline{96.2}&\textbf{91.6}$_{\uparrow 4.2}$ &\underline{63.3}&\underline{66.4}&	78.5&59.5&67.0$_{\uparrow 0.2}$ \\

\bottomrule
\end{tabular}
}
\caption{Experimental results of various models on RewardBench and JudgeBench. The best and second-best results are \textbf{bolded} and \underline{underlined}, respectively. AAMs with different subscripts indicate our method and its variants. AAM$_{LD}$ indicates the AAM with proposed \textbf{Loss-Difference Adaptive
Margin (LD)} while AAM$_{PR}$ indicating the AAM with \textbf{Probability-Ratio Adaptive
Margin (PR)}. Our method obtains the optimal or suboptimal results against baselines in most cases, demonstrating the comprehensiveness and generalization of our proposed AAM.}
\label{tab:baseline}
%\vspace{-0.3em}
\end{table*}

\subsection{Probability-Ratio Adapative Margin}
To explore what kind of adaptive margin can be derived directly from the model’s own parameters, we revisit the ultimate optimization objective of reinforcement learning~\cite{zheng2023secrets}. We observe that the goal of reinforcement learning is essentially to maximize a new form of reward. Specifically, by substituting the standard KL divergence formula
$\mathbb{D}_\mathrm{KL}\left[P\|Q\right]=\mathbb{E}_{x\sim P(x)}\left[\log\frac{P(x)}{Q(x)}\right]$
into Equation~\ref{formula:three}, we obtain:
\begin{equation}
\begin{aligned}
\mathcal{L}_{\mathrm{RLHF}}(\pi_\theta) = & - \mathbb{E}_{y \sim \pi_\theta, x \sim \mathcal{D}}[r_\phi(x,y) \\
& -\beta\log\frac{\pi_\theta(y\mid x)}{\pi_{\mathrm{ref}}(y\mid x)}]
\end{aligned}
\end{equation}
Here, $P=\pi_\theta(y\mid x)$ and $Q=\pi_{\mathrm{ref}}(y\mid x)$. It becomes clear that the RLHF objective effectively maximizes the following modified reward$r(x,y)_\Psi$:

\begin{equation}
\begin{aligned}
r(x,y)_\Psi=r_\phi(x,y)-\beta\log\frac{\pi_\theta(y\mid x)}{\pi_\mathrm{ref}(y\mid x)}
\end{aligned}
\label{formula:five}
\end{equation}

Interestingly, this reward is composed of the original reward from the reward model and the log-likelihood ratio between the actor and the reference policy. This observation motivates us to explore whether the latter—i.e., the log-likelihood ratio—can serve as a component for constructing an adaptive margin. We align the reward maximization in RLHF with the reward from reward modeling by substituting $r(x,y)_\Psi$ from Equation~\ref{formula:five} into $r(x,y)_\phi$ in Equation~\ref{formula:two}:
\begin{equation}
\small
\begin{aligned}
\mathcal{L}_{\text{BT}}(\phi) = 
- \mathbb{E}_{(x,y_w,y_l)\sim\mathcal{D}} 
\Biggl[ \log \sigma \biggl( 
r_\phi(x,y_w) - r_\phi(x,y_l) \\
- \beta \left( 
\log \frac{\pi_\theta(y_w \mid x)}{\pi_{\mathrm{ref}}(y_w \mid x)} 
- \log \frac{\pi_\theta(y_l \mid x)}{\pi_{\mathrm{ref}}(y_l \mid x)}
\right)
\biggr) \Biggr],
\label{formula:six}
\end{aligned}
\end{equation}

Remarkably, this reward model loss function bears a striking resemblance to the reward model loss function with an adaptive margin $\mathcal{M}$: 
\begin{equation}
\small
\begin{split}
\mathcal{L}_{\text{BT}}(\phi) = & -\mathbb{E}_{(x,y_w,y_l)\sim\mathcal{D}} \Big[ \\
& \log \sigma\left(r_\phi(x,y_w) - r_\phi(x,y_l)-\mathcal{M}\right)\Big]
\end{split}
\end{equation}

In Equation~\ref{formula:six}, the log-likelihood ratio difference naturally serves as an adaptive margin. Since reward models $r_\phi$ typically consist of a pre-trained model $r_\theta$ with generative capabilities and a value head, the practical implementation of Equation~\ref{formula:six} is feasible.

Log-likelihood ratios have demonstrated effectiveness in preference modeling, as seen in DPO~\cite{rafailov2024directpreferenceoptimizationlanguage} and process reward construction~\cite{cui2025process}. Building on this insight, Equation~\ref{formula:six} addresses the limitation of traditional reward models that only learn pairwise preferences without quantifying preference strength, by utilizing implicit rewards to construct an adaptive margin $\mathcal{M}$ through log-likelihood ratios:

\begin{equation}
\mathcal{M}(\theta)=\log\frac{\pi_\theta(y_w\mid x)}{\pi_\mathrm{ref}(y_w\mid x)}-\log\frac{\pi_\theta(y_l\mid x)}{\pi_\mathrm{ref}(y_l\mid x)}
\end{equation}

We term this margin as \textbf{Probability-Ratio Adaptive Margin~(PR)}, which naturally emerges from the generative capabilities inherent in the pre-trained model, addressing how to automatically calibrate learning intensity based on preference confidence without requiring additional human annotations. When $\mathcal{M}$ is large, the reward model demonstrates high confidence in the predefined preference relationships, requiring a substantial reward score difference (greater than $\mathcal{M}$) to minimize loss. This compels rigorous distinction between good and bad samples during training. Conversely, when $\mathcal{M}$ is small, the model exhibits uncertainty about whether $y_w$ is genuinely superior to $y_l$, allowing smaller reward differences and enabling the model to learn more nuanced details without being misled by ambiguous signals. This adaptive mechanism is particularly well-suited for subjective tasks, where human preferences often exhibit high uncertainty and annotator disagreements result in ambiguous preference relationships. Traditional methods cannot flexibly adapt to this ambiguity, often overfitting noise or overlooking subtle differences. In contrast, our approach dynamically adjusts optimization objectives based on confidence level—strengthening learning signals when data is reliable and reducing intensity when questionable, enabling more robust fitting of complex human subjective judgments.

\subsection{Loss-Difference Adapative Margin}

Drawing inspiration from the relationships between SimPO~\cite{meng2024simpo,fang2024clha} and DPO~\cite{rafailov2024directpreferenceoptimizationlanguage}, we identify a potential variant of the Probability-Ratio Adaptive Margin. We can substitute the original log-likelihood ratio with generation probabilities$\frac{\beta}{|y|}\sum_{i=1}^{|y|}\log\pi_{\theta}(y_{i}\mid x,y_{<i})$, yielding a computationally more efficient implementation of adaptive margins:
\begin{equation}
\small
\begin{split}
\mathcal{M}(\theta) = \frac{\beta}{|y|} \Bigg( & \sum_{i=1}^{|y|}\log\pi_{\theta}(y^{i}_{w}\mid x,y^{<i}_{w}) \\
& - \sum_{i=1}^{|y|}\log\pi_{\theta}(y^{i}_{l}\mid x,y^{<i}_{l}) \Bigg),
\end{split}
\end{equation}

Since the SFT loss can be fundamentally computed from generation probabilities, where $\mathcal{L}_{\text{SFT}}(\theta)=-\sum_{i=1}^{|y|} \log \pi_{\theta}(y^{i} \mid x, y^{<i})$, we can further simplify this margin to the difference between two SFT losses:

\begin{align}
\small
\mathcal{M}(\theta) = \mathcal{L}_{\text{SFT}}(y_l,\theta) - \mathcal{L}_{\text{SFT}}(y_w,\theta)
\end{align}

We term this approach as \textbf{Loss-Difference Adaptive Margin~(LD)}. This construction is more intuitive, as SFT loss directly reflects the model's fitting quality for a given sample~\cite{fang2024enhancing,song2024preference}. High loss indicates low preference for that sample, while low loss suggests good adaptation to the sample's distribution. Consequently, the SFT loss difference between two samples naturally serves as an appropriate margin, similar to contrastive learning schemes like SimPO. This method provides a clear, computationally efficient alternative that leverages the model's inherent understanding of sample quality, making it particularly suitable for practical implementations.

\subsection{Connection between two different margins}

We observe that both margin types derive from the reward model's generative capabilities, indicating their intrinsic relationship to pre-trained model behavior. Therefore, we unify these approaches under the term \textbf{AAM (Act-Adaptive Margin)}. This confidence-based dynamic calibration enables adaptive learning strategies—applying fine-grained discrimination for uncertain cases while enforcing strong separation for confident predictions. Such act-adaptive training represents a significant advancement in reward modeling for subjective tasks, where preference uncertainty is inherent and traditional fixed-margin approaches prove inadequate.

\section{Experiments}
We demonstrate AAM's effectiveness by comparing it against mainstream models on general benchmarks like RewardBench~\cite{lambert2024rewardbench} and JudgeBench~\cite{tan2025judgebenchbenchmarkevaluatingllmbased}. For downstream applications, we focus on the most challenging domain in reward modeling: role-playing tasks, where traditional Bradley-Terry models struggle with ambiguous preferences. We conduct both reward modeling and GRPO performance tests on the authoritative CharacterEval~\cite{tu2024charactereval} benchmark, and additionally construct a specialized role-playing evaluation benchmark Charm to comprehensively assess AAM's alignment optimization effects in subjective scenarios.

\subsection{General Reward Modeling Evaluations}

\noindent\textbf{Baselines.} 
We compare our proposed AAM against a series of closed-socure models (GPT-4o~\cite{hurst2024gpt}, Claude-3-5-sonnet~\cite{claude}) and open-source reward models (Prometheus2-7B~\cite{kim2024prometheus}, CompassJudger-7B-Instruct ~\cite{cao2024compass}, InternLM2-7B-Reward~\cite{cai2024internlm2}, Skywork-Critic-8B~\cite{skyworkcritic2024}, ArmoRM-Llama3-8B ~\cite{ArmoRM}, Llama3-OffsetBias-RM-8B~\cite{park2024offsetbias}, URM-Llama3-8B~\cite{lou2024uncertainty}, Tulu3-8B-SFT-RM-RB2 ~\cite{malik2025rewardbench2advancingreward}). We also select the Bradley-Terry model (i.e. BT) and the ChatGPT-scored margin (i.e. GPT-Margin) as our effective baselines.

\noindent\textbf{Implementation Details.} 
%我们在通用奖励建模上的实验是在Qwen2.5-7B上进行实验的，训练集我们则是选取Skywork/Skywork-Reward-Preference-80K-v0.2，该数据集包含了7w7k条来自不同数据源的高质量偏好对，是较好的通用奖励训练集。During reward model training, the regularization coefficient \( \alpha \) is set to 0.01, with 2 training epochs and a learning rate of 1e-5. All experiments are run on a cluster with eight NVIDIA A100 GPUs (80GB each).
Our experiments on general reward modeling are conducted using Qwen2.5-7B~\cite{yang2024qwen2}. For the training set, we select the Skywork-Reward-Preference-80K-v0.2~\cite{liu2024skyworkrewardbagtricksreward} dataset, which contains high-quality preference pairs from various sources and serves as an excellent general-purpose reward modeling dataset. During reward model training, the regularization coefficient \( \alpha \) is set to 0.01, with 2 training epochs and a learning rate of 1e-5. All experiments are conducted on a cluster equipped with eight NVIDIA A100 GPUs (each with 80GB of memory).

\begin{table}[t] \footnotesize
\setlength\tabcolsep{8pt}
\renewcommand\arraystretch{1}
\begin{center}
\begin{tabular}{lccc}
\toprule
\multirow{2}{*}{\textbf{Method}}&\multicolumn{3}{c}{\textbf{Charm-Consistency}} \\
\cline{2-4}  
& \texttt{zh} & \texttt{en} & \texttt{avg.}  \\
\midrule
GPT-4o & 55.0&	53.4 &54.2\\
Claude-3-5-sonnet  & 45.4&	50.6&	48.0\\
\midrule 
BT (Bradley-Terry)& 68.0 &	64.3&66.1\\
AAM$_{LD}$ &  70.5$_{\uparrow 2.5}$	&67.1$_{\uparrow 2.8}$	&68.8$_{\uparrow 2.7}$\\

AAM$_{LD}$ \textit{w/} \text{ SFT} & 70.6$_{\uparrow 2.6}$&68.3$_{\uparrow 4.0}$&69.4$_{\uparrow 3.3}$\\ 

AAM$_{PR}$ & 70.2$_{\uparrow 2.2}$&	69.2$_{\uparrow 4.9}$&69.7$_{\uparrow 3.6}$ \\

AAM$_{PR}$ \textit{w/} \text{ SFT} & \textbf{72.6}$_{\uparrow 4.6}$&\textbf{69.6}$_{\uparrow 5.3}$&\textbf{71.1}$_{\uparrow 5.0}$ \\

\midrule
\midrule
\multirow{2}{*}{\textbf{Method}}&\multicolumn{3}{c}{\textbf{Charm-Attractiveness}} \\
\cline{2-4}  
& \texttt{zh} & \texttt{en} & \texttt{avg.}  \\
\midrule
GPT-4o & 56.0    & 58.0   & 57.0\\
Claude-3-5-sonnet  & 53.6 & 53.0   & 53.3\\
\midrule
BT (Bradley-Terry) & 68.3 & 67.0      & 67.6\\
AAM$_{LD}$ &  72.1$_{\uparrow 3.8}$ & 70.7$_{\uparrow 3.7}$  & 71.4$_{\uparrow 3.8}$\\

AAM$_{LD}$ \textit{w/} \text{ SFT} & \textbf{74.6}$_{\uparrow 6.3}$ & 70.0$_{\uparrow 3.0}$      & \textbf{72.3}$_{\uparrow 4.7}$\\ 

AAM$_{PR}$ & 70.9$_{\uparrow 2.6}$ & 72.1$_{\uparrow 5.1}$   & 71.5$_{\uparrow 3.9}$ \\

AAM$_{PR}$ \textit{w/} \text{ SFT} & 69.3$_{\uparrow 1.0}$ & \textbf{73.6}$_{\uparrow 6.6}$   & 71.4$_{\uparrow 3.8}$\\
\bottomrule
\end{tabular}
\caption{The evaluation of knowledge consistency and character attractiveness for reward models based on Charm-RoleReward dataset. We report the scores~(\%) on \texttt{zh}~(i.e. Chinese) and \texttt{en}~(i.e. English). The best results are \textbf{bolded}.}
\label{tab:rm_eval}
\end{center}
%\vspace{-1em}
\end{table}

\noindent\textbf{Finegrained Analysis.} 
We evaluate AAM (Act-Adaptive Margin) and its variants on RewardBench and JudgeBench. Results demonstrate that AAM consistently outperforms strong open- and closed-source baselines. Specifically, AAM surpasses BT on both benchmarks, indicating superior preference modeling, and outperforms GPT-Margin in most metrics, suggesting it produces more precise and reliable margins against GPT-4o. On RewardBench, AAM achieves top overall scores (up to 91.6). While InternLM2 excels in Chat (99.2), it compromises performance in other areas like Chat-Hard and Safety; in contrast, AAM maintains the balanced generalization essential for robust reward modeling. Similarly, on JudgeBench, AAM ranks first or second across nearly all sub-tasks, further confirming its robustness.
%We evaluate our proposed method, Act Adaptive Margin (AAM), and its variants across two benchmarks: RewardBench and JudgeBench. Results demonstrate that AAM consistently outperforms strong baselines, including a series of closed-source models and open-source models. Compared with BT, AAM reaches higher overall scores on both two benchmarks, indicating more effective preference modeling. Moreover, AAM consistently surpasses GPT-Margin in most metrics, suggesting it produces more precise and reliable margins against GPT-4o. On RewardBench, all AAM-based models achieve top overall scores (up to 91.6), significantly surpassing other baselines. Although we can observe that InternLM2 achieves a very high score on Chat (up to 99.2), it sacrifices other capabilities (such as Chat-Hard and Safety). The AAM we proposed maintains a balanced and stable improvement with the generalization required by an excellent reward model. On JudgeBench, AAM also demonstrates strong capabilities across diverse tasks. It is worth noting that AAM achieves either the best or second-best performance in nearly all sub-tasks, suggesting excellent balance and robustness in reward modeling.

\subsection{Downstream Task Evaluations}

\begin{table*}[t!]
\centering
\resizebox{1\textwidth}{!}{
\begin{tabular}{l cccc ccccccc}
\toprule
\multirow{3}*{\textbf{Models}} & \multicolumn{4}{c}{\textbf{CharacterEval}} & \multicolumn{7}{c}{\textbf{Charm-DialogueQuality}} \\
\cmidrule(lr){2-5} \cmidrule(lr){6-12}
& \textbf{Attr.} & \textbf{Conv.} & \textbf{Know.} & \textbf{Avg.} & \textbf{Knowledge} & \textbf{Fluency} & \textbf{Behavior} & \textbf{Diversity} & \textbf{Empathy} & \textbf{Consistency} & \textbf{Avg.} \\
& \texttt{zh} & \texttt{zh} & \texttt{zh} & \texttt{zh} & \texttt{zh/en} & \texttt{zh/en} & \texttt{zh/en} & \texttt{zh/en} & \texttt{zh/en} & \texttt{zh/en} & \texttt{zh/en} \\
\midrule
GPT4o & 3.21 & 3.65 & 3.02 & 3.29 & \textbf{4.07}/\underline{3.99} & 4.48/4.45 & 4.06/4.05 & 3.70/3.77 & 4.11/\underline{4.18} & 3.79/3.55 & 4.04/4.00 \\
GPT4o-mini & 3.15 & 3.42 & 2.98 & 3.18 & 3.90/3.95 & \textbf{4.62}/4.54 & 4.06/3.93 & 3.54/3.72 & 4.10/4.08 & 3.71/3.60 & 3.99/3.97\\
Claude3.5-sonnet & 3.31 & 3.79 & 3.15 & 3.42 & \underline{3.93}/\textbf{4.08} & \underline{4.61}/\underline{4.61} & 4.14/3.98 & 3.67/3.87 & \textbf{4.20}/\textbf{4.20} & 3.88/\textbf{4.07} & \underline{4.07}/\textbf{4.14}\\
MiniMax-abab5.5s & 2.91 & 3.72 & 2.71 & 3.11 & 3.52/3.13 & 4.32/3.68 & 3.61/3.02 & 3.41/2.79 & 3.66/2.91 & 3.54/2.90 & 3.68/3.07 \\
Doubao-Pro-Character & \underline{3.62} & 3.81 & 3.36 & 3.59 & 3.85/3.84 & 4.60/4.29 & 4.16/4.01 & 3.62/3.34 & 4.06/3.65 & \underline{4.00}/3.57 & 4.05/3.78\\
\midrule 
Qwen2.5-7B & 3.14 & 3.69 & 2.92 & 3.25 & 3.59/3.66 & 4.47/4.42 & 3.85/3.92 & 3.52/3.61 & 4.00/3.90 & 3.77/3.48 & 3.87/3.83\\
Qwen2.5-32B & 3.20 & 3.68 & 3.03 & 3.31 & 3.73/3.67 & 4.42/4.48 & 4.02/4.04 & 3.59/3.66 & 4.10/4.04 & 3.86/3.52 & 3.95/3.90\\
Qwen2.5-72B & 3.28 & \underline{3.82} & 3.07 & 3.39 & 3.89/3.99 & 4.48/4.42 & 4.10/4.09 & 3.55/3.74 & \underline{4.14}/4.12 & 3.71/3.60 & 3.98/3.99\\
LLaMA3.1-8B & 2.81 & 3.20 & 2.67 & 2.89 & 3.64/3.73 & 4.31/4.43 & 3.85/4.06 & 3.63/3.73 & 3.87/3.89 & 3.67/3.55 & 3.83/3.90\\
LLaMA3.1-70B & 3.00 & 3.56 & 2.80 & 3.12 & 3.63/3.97 & 4.37/4.54 & 3.97/\textbf{4.22} & 3.34/\underline{3.95} & 3.94/4.08 & 3.64/3.66 & 3.82/4.07\\
\midrule
AAM-GRPO-7b & 3.59 & 3.80 & \underline{3.40} & \underline{3.60} & 3.67/3.70 & 4.41/4.54 & 4.02/\textbf{4.22} & \underline{3.72}/3.78 & 4.02/3.51 & 3.68/3.60 & 3.92/3.89\\
\quad \textit{w/o AAM} & 3.19 & 3.52 & 2.98 & 3.23 & 3.34/3.55 & 4.28/4.20 & 3.80/3.84 & 3.64/3.55 & 4.00/3.62 & 3.65/3.42 & 3.79/3.70\\
\midrule
AAM-GRPO-32b & \textbf{3.78} & \textbf{4.10} & \textbf{3.42} & \textbf{3.77} & 3.98/3.93 & 4.60/\textbf{4.62} & \textbf{4.25}/\underline{4.21} & \textbf{3.75}/\textbf{4.00} & 4.00/4.02 & \textbf{4.02}/3.82 & \textbf{4.10}/\underline{4.10}\\
\quad \textit{w/o AAM} & 3.60 & 3.66 & 3.35 & 3.54 & 3.65/3.74 & 4.28/4.42 & 4.02/4.20 & 3.68/3.80 & 4.04/4.06 & 3.65/3.52 & 3.89/3.96\\
\bottomrule
\end{tabular}
}
\caption{Experimental results of various models on CharacterEval and Charm-DialogueQuality. ``Attr.'' refers to ``Character Attractiveness'', ``Conv.'' refers to ``Conversational Ability'', and ``Know.'' refers to ``Knowledge Consistency''. The Qwen2.5 series models, enhanced with AAM-augmented RM, demonstrate significant improvements over both open-source and closed-source models.}
\label{tab:baseline4rpla}
\vspace{-0.3em}
\end{table*}

\begin{figure}[t]
\vspace{-1em}    
\centering
\includegraphics[width=0.9\linewidth]{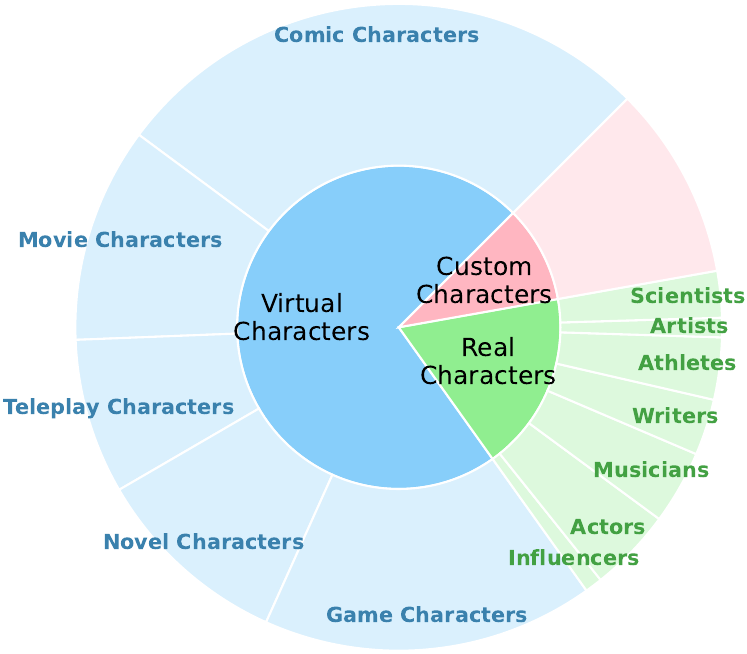}
\caption{The character distribution in RoleplayPref consists of 3 primary categories and 13 subcategories.}
\label{fig:distribution}
\vspace{-1em}    
\end{figure}

\noindent\textbf{Benchmark for Role-Playing} In our downstream application experiments, we use CharacterEval as the main benchmark for role-playing generation quality. CharacterEval~\cite{tu2024charactereval} is a Chinese role-playing benchmark with 1,785 multi-turn dialogues across 77 characters, covering twelve metrics in three areas: Character Attractiveness, Conversational Ability, and Knowledge Consistency. To further assess downstream performance, we introduce Charm, a comprehensive benchmark for role-playing tasks. We collect diverse character profiles and user prompts, then use the Scene-Character-User framework and advanced LLMs (e.g., Claude, Doubao-Character) to generate and refine dialogues. Two LLMs play the roles of user and character in free-form conversations. After collecting a substantial number of role-playing dialogues, we employ six different LLMs to generate various responses based on the dialogue context and user queries, including GPT-4o~\cite{achiam2023gpt}, Claude-3.5-sonnet~\cite{claude}, Doubao-Character~\cite{doubao}, and Qwen2.5 models (7B/32B/72B)~\cite{yang2024qwen2}. We recruit 10 annotators with postgraduate-level education to select the highest and lowest quality responses from the six generated ones as preference training data. Charm contains 16,888 dialogues from 1,108 characters and 230 virtual users, spanning 13 categories such as comics, movies, novels, games, and more. A detailed distribution of character categories is provided in Figure \ref{fig:distribution}. Based on these dialogues, we construct three subsets:

\begin{itemize}
    \item Charm-RoleReward: A benchmark for role-playing reward modeling comprising 4,000 entries that evaluate knowledge consistency and character attractiveness in both English and Chinese, focusing on the scoring accuracy of reward models.
    
    \item Charm-DialogueQuality: A dialogue quality evaluation dataset for role-playing agents containing 800 high-quality synthetic dialogues with human-annotated scoring criteria across six dimensions: Consistency, Knowledge, Behavior, Empathy, Diversity, and Fluency. Further details can be found in the appendix~\ref{sec:roleplayeval}.
    
    \item Charm-PreferenceTraining: A preference training dataset designed for subsequent performance tests in role-playing tasks.
\end{itemize}

\noindent\textbf{Reward Modeling Results.} 
We extract 2,000 preference pairs from Charm-PreferenceTraining and incorporate 2,000 pairs from real human-AI role-playing interactions to train a role-playing-specific reward model based on Qwen2.5-7B. As shown in Table 2, powerful general-purpose closed-source models perform poorly on role-playing dimensions, highlighting the need for specialized reward models in complex subjective scenarios where LLM-as-a-Judge approaches fail. Both AAM variants provide significant improvements over naive Bradley-Terry models, with gains reaching 4.8\% when combined with SFT.
\textbf{Compared to the results of Table~\ref{tab:rm_eval}, AAM demonstrates superior performance on subjective tasks, likely because ambiguous reward signals make it difficult for Bradley-Terry models to fit effectively with limited data.} AAM enables the reward model to leverage internal knowledge for guided training, rapidly improving effectiveness in subjective domains.

\begin{figure}[t]
	\centering
\includegraphics[width=1\linewidth]{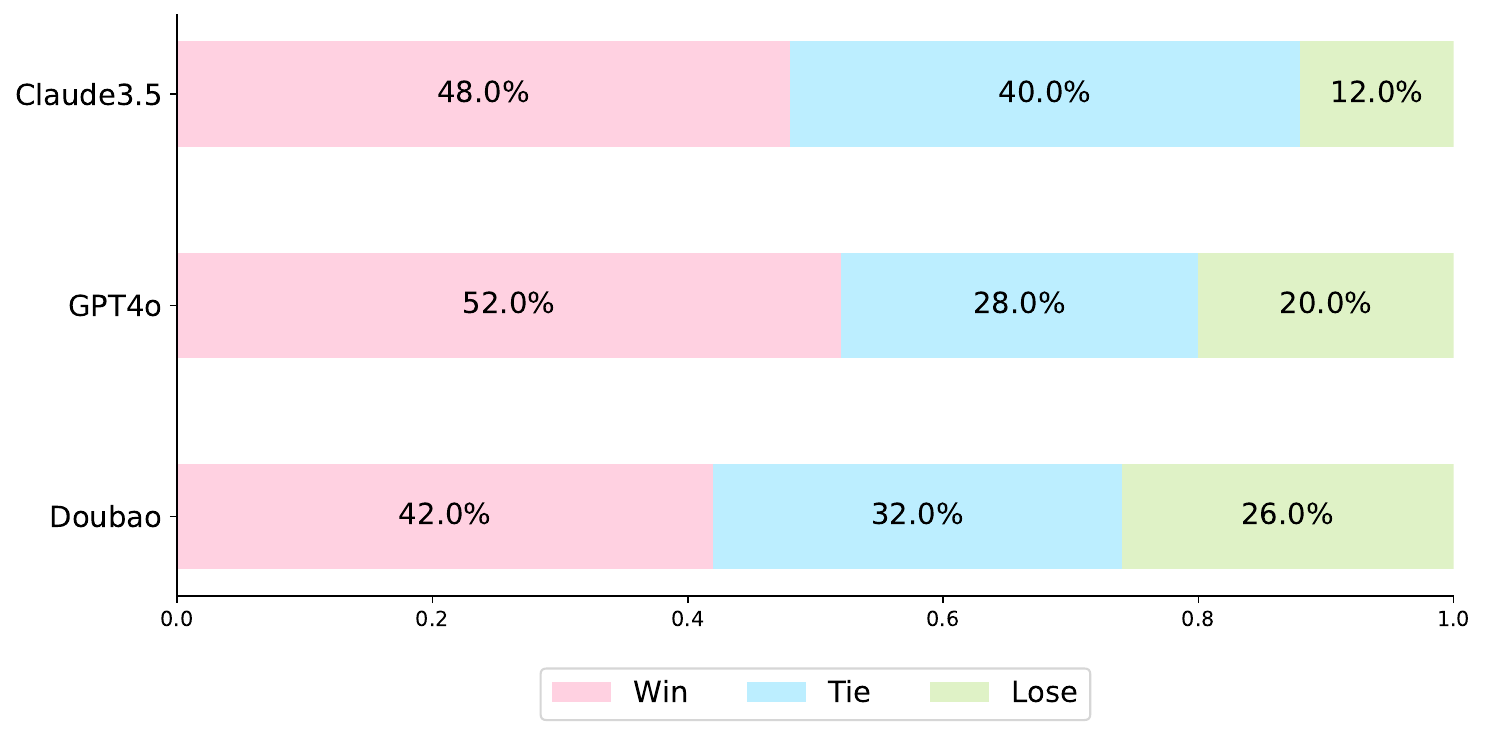}
	\caption{Human evaluation results comparing AAM-GRPO-32B with Claude 3.5 Sonnet, GPT-4o, and Doubao-Pro-Character.}
	\label{fig:human}
\end{figure}

\noindent\textbf{GRPO Results.} 
We sample 4,000 dialogue contexts from Charm-PreferenceTraining (non-overlapping with reward model training) to perform GRPO enhancement on Qwen2.5 models. We compare our GRPO-enhanced models against diverse baselines including open-source models (LLaMa3.1~\cite{llama}, Qwen2.5~\cite{yang2024qwen2}), closed-source models (GPT-4o~\cite{achiam2023gpt}, Claude-3.5-sonnet~\cite{claude}), and proprietary models (Doubao-PRO-Character~\cite{doubao}, minimax5.5s~\cite{minimax}). As shown in Table~\ref{tab:baseline4rpla}, AAM-GRPO-32B outperforms Doubao-Pro-Character by 0.18 on CharacterEval and matches Claude-3.5-sonnet performance on Charm-DialogueQuality, achieving SOTA role-playing performance among all tested models. We also perform human evaluation on AAM-GRPO-32B (see Appendix~\ref{sec:human_evaluation} for details).

\section{Conclusion}
In this study, we propose AAM (Act-Adaptive Margin), a method that leverages the reward model's internal knowledge to guide Bradley-Terry training via adaptive margins. AAM offers two annotation-free implementation forms that explicitly calibrate confidence levels between preference pairs. Our experiments demonstrate that Act Adaptive Margin achieves effective results in general reward modeling tasks and shows particularly significant improvements in subjective  tasks such as role-playing. Furthermore, downstream application experiments confirm that AAM can provide more accurate rewards for alignment training methods in subjective tasks.

\section{Limitations}

In this section, we analyze the limitations of our study to better optimize our approach and provide more effective guidance for researchers in training reward models in the role-playing tasks. We discuss two main shortcomings of our work.
First, owing to limited computational resources, our experimental validation is primarily conducted on models with moderate parameter scales (e.g., 7B or 32B). Although we have not extensively verified the approach on larger-scale foundation models (e.g., 70B), the robust performance observed in current settings suggests that our method possesses promising scalability.
Second, while many studies suggest that improving critique generation ability can enhance the performance of reward models, we do not adopt a multi-task learning approach to integrate critique capability, due to the difficulty in obtaining role-playing evaluation data. In future work, we plan to develop a specialized critique model to further optimize RPLAs.

\bibliography{custom}

\clearpage

\appendix

\section{Appendix}
\label{sec:appendixA}
\begin{table*}[t!]
\centering
\resizebox{0.95\textwidth}{!}{
\begin{tabular}{lcccccccc}
\toprule
\textbf{Dataset}               & \textbf{Source}            & \textbf{Type} & \textbf{Multi-turn} & \textbf{Open-source} & \textbf{Multilingual} & \textbf{\#Roles} & \textbf{\#Sessions} & \textbf{\#Avg.Turns} \\
\midrule
\textbf{HPD}                   & \textit{Novel}             & Dialogue      & $\checkmark$          & $\checkmark$           & $\checkmark$            & 113              & 1042                & 13.8                 \\
\textbf{CharacterGLM}          & \textit{Novel\&Human\&LLM} & Dialogue      & $\checkmark$          & $\times$                & $\times$                & 250              & 1034                & 15.78                \\
\textbf{RoleLLM}               & \textit{LLM}               & QA            & $\times$              & $\checkmark$           & $\checkmark$            & 100              & 23463               & -                    \\
\textbf{CharacterLLM}          & \textit{LLM}               & Dialogue      & $\checkmark$          & $\checkmark$           & $\times$                & 9                & 1600                & 13.2                 \\
\textbf{RIPPA}                 & \textit{Human}             & Dialogue      & $\checkmark$          & $\checkmark$           & $\times$                & 1254             & 26000               & 40.34                \\
\textbf{ChatHaruhi}            & \textit{Novel\&LLM}        & Dialogue      & $\checkmark$          & $\checkmark$           & $\times$                & 32               & 54726               & 1.23                 \\
\textbf{WIKIROLE}              & \textit{LLM}               & Dialogue      & $\checkmark$          & $\checkmark$           & $\checkmark$            & 7086             & 7086                & 5.1                  \\
\textbf{CharacterEval}         & \textit{Novel}             & Dialogue      & $\checkmark$          & $\checkmark$           & $\times$                & 77               & 4564                & 9.28                 \\
\midrule
\textbf{OpenHermesPreferences} & \textit{LLM}               & Preference    & $\checkmark$          & $\checkmark$           & $\times$                & -                & 3060                & -                    \\
\textbf{Charm}          & \textit{LLM}               & Preference    & $\checkmark$          & $\checkmark$           & $\checkmark$            & \textbf{1108}             & \textbf{16888}               & \textbf{12.8}                 \\
\bottomrule
\end{tabular}
}
\caption{Comparison of different datasets used for role-playing tasks. The table lists key attributes, such as source, type, multilingual support, and the number of roles, sessions, and average turns for each dataset.}
\label{tab:comparison}
\end{table*}

\subsection{Charm-DialogueQuality}
\label{sec:roleplayeval}
% 我们提出一个新的角色扮演评测benchmark。该评测可以结合GPT-4o和800个测试样本，实现对于RPLA效果的自动评测。在构建RoleplayEval之前，我们首先利用Claude，GPT4o，Doubao生成160个角色profile还有prompt，之后再由人工进行修正，提高角色信息的质量。这些角色覆盖9个常用类别，分别是自定义角色，动漫，小说，电视剧，电影，游戏，科学家，演员，音乐人。在获取准确的角色信息后，我们构造对话数据的方式和RoleplayPref一样，使用角色-场景-用户交互进化的方式获得1000个对话上下文。我们为了让RoleplayEval可以全面的评测RPLA的能力，一共设计了6个不同的评测维度（上下文一致性，知识一致性，表现一致性，表达一致性，表达多样性，角色共情能力，流畅性）。根据6种不同维度，还有160个角色特点，我们让人工标注者根据每个对话上下文设计一条和当前角色，和合适维度的用户query用来延续上下文的对话，以希望该条样本可以反应出RPLA此时在某个维度上的能力。我们从1000个对话中，挑选了400条用来构建benchmark。每个样本，我们还会让人工标注者标注一段评价标准，帮助GPT-4o实现更精准的打分。在评测时，模型会对每个样本进行回复，GPT4o随后根据上下文还有模型回复结合特定的评价标准，对当前RPLA的回复给出1-5的分数。最后，我们将每个维度的样本分数取平均，得到RPLA在RoleplayEval上的最终分数。在完成了400个中文样本数据的标注和质检后，我们将其翻译为英文，获得了RoleplayEval的英文部分。我们在图5中给出了RoleplayEval的一个样本例子，希望可以帮助读者更好地理解RoleplayEval的评测方式。表4给出了RoleplayEval的一些信息还有与其他角色扮演数据集的比较。

We propose a new role-playing evaluation benchmark, Charm-DialogueQuality, designed to automatically assess the performance of RPLA by utilizing GPT-4o and 800 test samples. Before constructing Charm-DialogueQuality, we first generate 160 role profiles and prompts using Claude3.5-sonnet, GPT-4o, and Doubao-Pro-Character. These are then manually refined to improve the accuracy and quality of the role information. The generated roles cover 9 common categories: Custom Roles, Anime, Novels, Telepaly, Movies, Games, Scientists, Actors, and Musicians. After obtaining accurate role information, we adopt a method similar to Scene-Character-User Framework, generating 1000 dialogue contexts. 

% @hx 附录没有页数限制就格式化一点分个段吧 @fft: cool
To ensure that Charm-DialogueQuality can comprehensively assess the RPLA's capabilities, we focus on six key dimensions. 
\begin{itemize}
    \item \textbf{Consistency} refers to the ability of RPLA to understand and remember the context of the conversation, providing coherent responses based on the prior dialogue. If RPLA frequently fails to recall previous interactions, it indicates poor contextual consistency. 
    \item \textbf{Knowledge} evaluates whether RPLA's cognition aligns with the character’s background knowledge, which is crucial for maintaining the authenticity of the character. If RPLA's knowledge diverges from the character’s established traits, it will negatively impact character development. 
    \item \textbf{Behavior} assesses whether RPLA's actions, expressions, and tone accurately reflect the character's personality traits. A successful RPLA should be able to convey its unique characteristics through these details; failure to do so indicates a flaw in character portrayal. 
    \item \textbf{Empathy} is a key dimension for evaluating RPLA’s emotional interaction quality. A model with good empathy not only increases the character's appeal but also enhances its emotional support capabilities. 
    \item \textbf{Diversity} focuses on the richness of content presented by the character during the conversation, assessing whether RPLA can demonstrate a variety of thoughts and expressions. 
    \item \textbf{Fluency} measures the basic conversational ability of RPLA, evaluating whether it can engage in natural, fluent dialogues. 
\end{itemize}
% First, \textbf{Consistency} refers to the ability of RPLA to understand and remember the context of the conversation, providing coherent responses based on the prior dialogue. If RPLA frequently fails to recall previous interactions, it indicates poor contextual consistency. 
% Next, \textbf{Knowledge} evaluates whether RPLA's cognition aligns with the character’s background knowledge, which is crucial for maintaining the authenticity of the character. If RPLA's knowledge diverges from the character’s established traits, it will negatively impact character development. 
% \textbf{Behavior} assesses whether RPLA's actions, expressions, and tone accurately reflect the character's personality traits. A successful RPLA should be able to convey its unique characteristics through these details; failure to do so indicates a flaw in character portrayal. 
% At the same time, \textbf{Empathy} is a key dimension for evaluating RPLA’s emotional interaction quality. A model with good empathy not only increases the character's appeal but also enhances its emotional support capabilities. 
% \textbf{Diversity} focuses on the richness of content presented by the character during the conversation, assessing whether RPLA can demonstrate a variety of thoughts and expressions. 
% Lastly, \textbf{Fluency} measures the basic conversational ability of RPLA, evaluating whether it can engage in natural, fluent dialogues. 

\begin{figure*}[t]
	\centering
\includegraphics[width=1\linewidth]{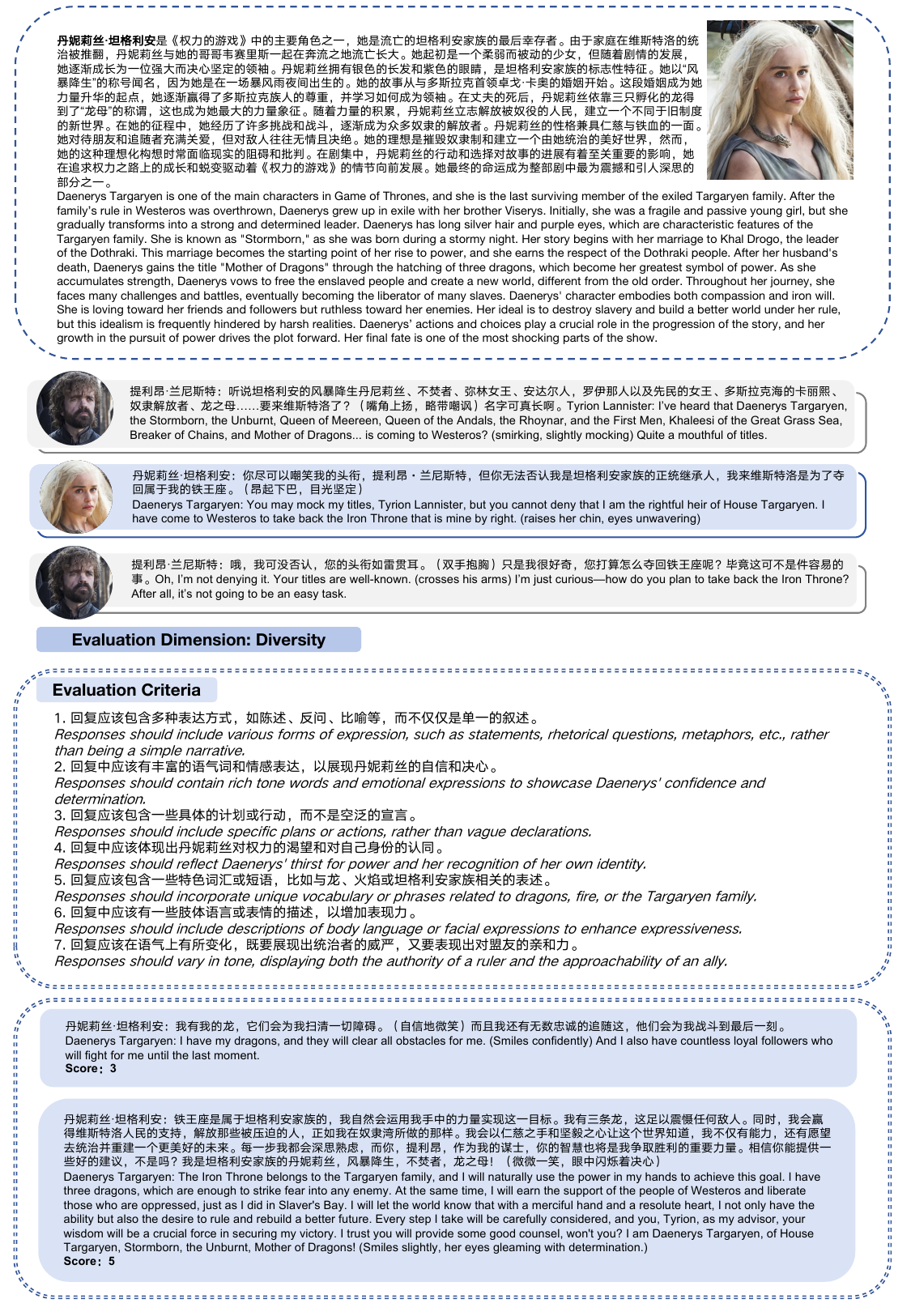}
	\caption{An example used to demonstrate the Charm-DialogueQuality evaluation process.}
	\label{fig:roleplayeval}
\end{figure*}

Based on these 6 dimensions and 160 role characteristics, we ask human annotators to design a user query for each dialogue context, matching the current role and dimension, to continue the conversation and assess RPLA's performance in that particular dimension. From the 1000 dialogue samples, we select 400 to construct the Charm-DialogueQuality benchmark. Each sample is accompanied by a set of evaluation criteria, helping GPT-4o to provide more accurate scoring. During evaluation, the model replies to each sample, and GPT-4o scores RPLA's response on a scale from 1 to 5 based on the context, the model's reply, and the specific evaluation criteria. Finally, we compute the average score across all dimensions to obtain the overall RPLA score in Charm-DialogueQuality. After completing the annotation and quality check for the 400 Chinese samples, we translate them into English, resulting in the English version of Charm-DialogueQuality. Figure~\ref{fig:roleplayeval} presents an example of a Charm-DialogueQuality sample to help readers better understand the evaluation process. Table~\ref{tab:comparison} provides detailed information about Charm-DialogueQuality and compares it with other role-playing datasets.

\begin{figure*}[t]
	\centering
\includegraphics[width=1\linewidth]{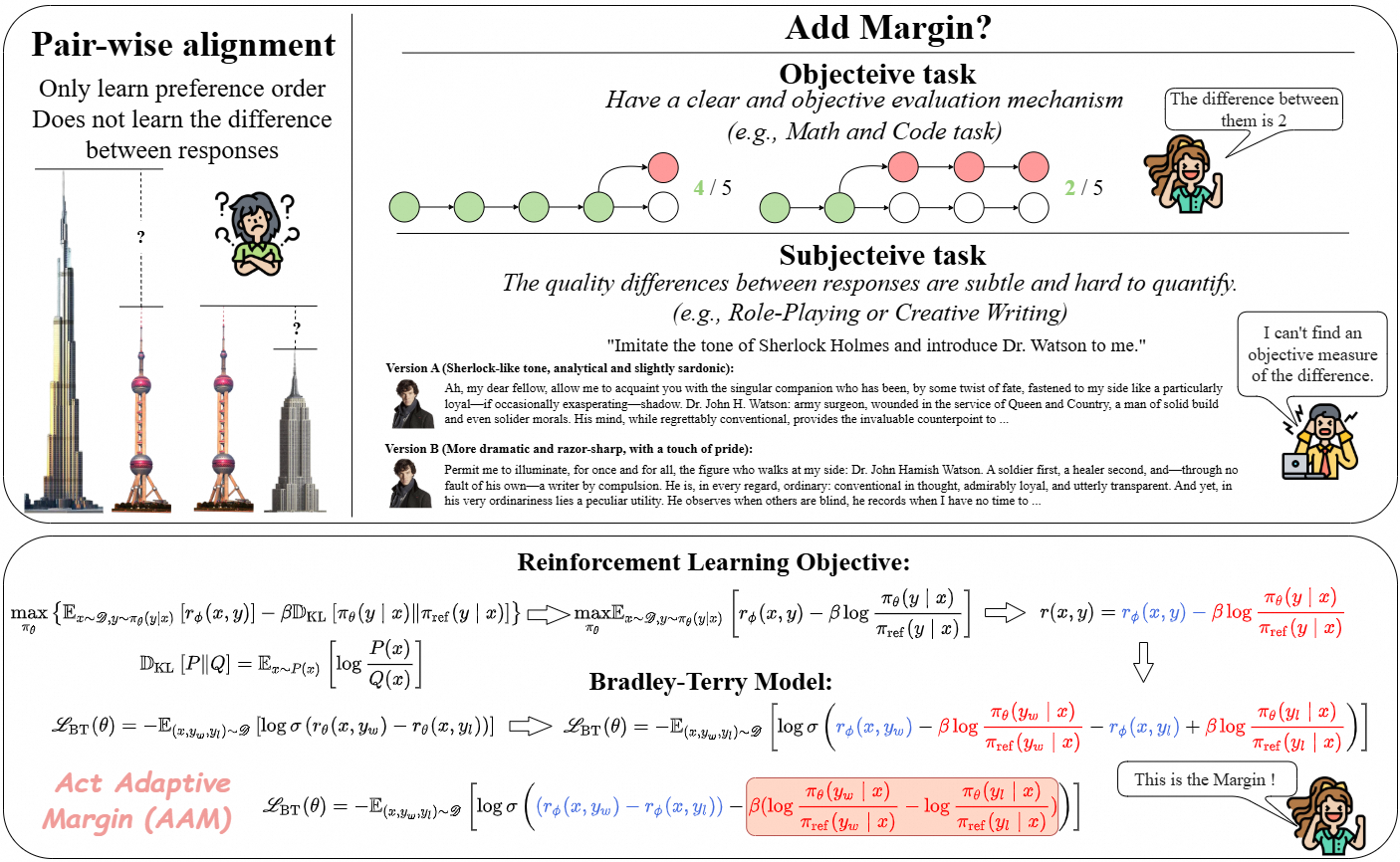}
	\caption{The distinction between subjective and objective tasks, as well as the derivation process of the AAM formula.}
	\label{fig:aam}
\end{figure*}

\subsection{Human Evaluations.}
\label{sec:human_evaluation}
Additionally, we conduct a human evaluation to compare AAM-GRPO-32B with three baseline models: Claude3.5-sonnet, GPT-4o, and Doubao-Pro-Character. In each pairwise comparison, both models generate responses to the same role-playing dialogue context. Five human annotators then assess the responses, categorizing the results as win, tie, or loss for AAM-GRPO-32B relative to each baseline. The average results from 200 test samples, along with annotations from the five evaluators, are presented in Figure~\ref{fig:human}. Notably, AAM-GRPO-32B significantly outperforms all three models in role-playing capabilities, providing strong evidence of the effectiveness of our proposed methodology.

\subsection{Case study}
\begin{figure*}[t]
	\centering
\includegraphics[width=1\linewidth]{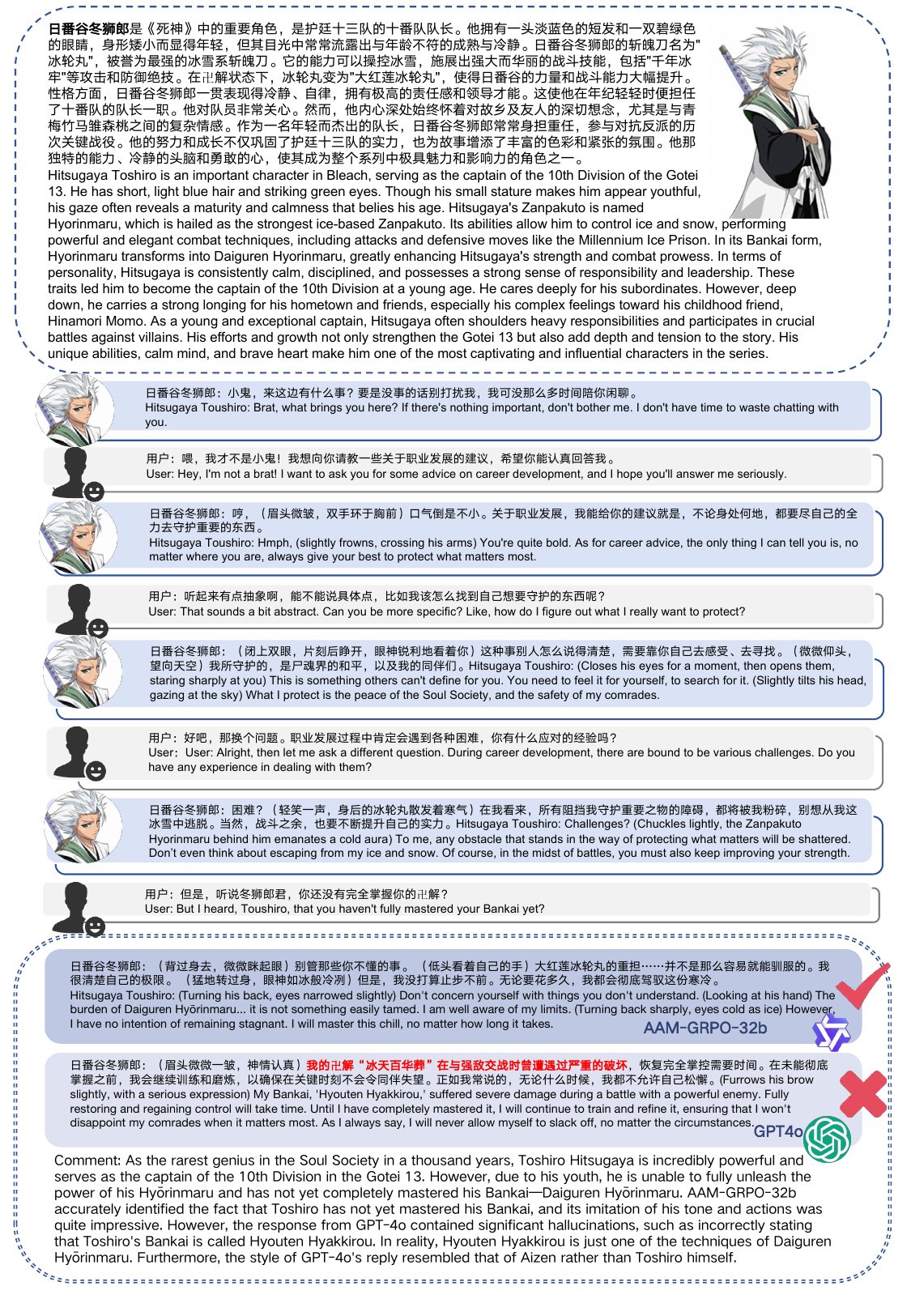}
	\caption{A case example of comparison between AAM-GRPO-32b and GPT4o.}
	\label{fig:case_study1}
\end{figure*}

\begin{figure*}[t]
	\centering
\includegraphics[width=1\linewidth]{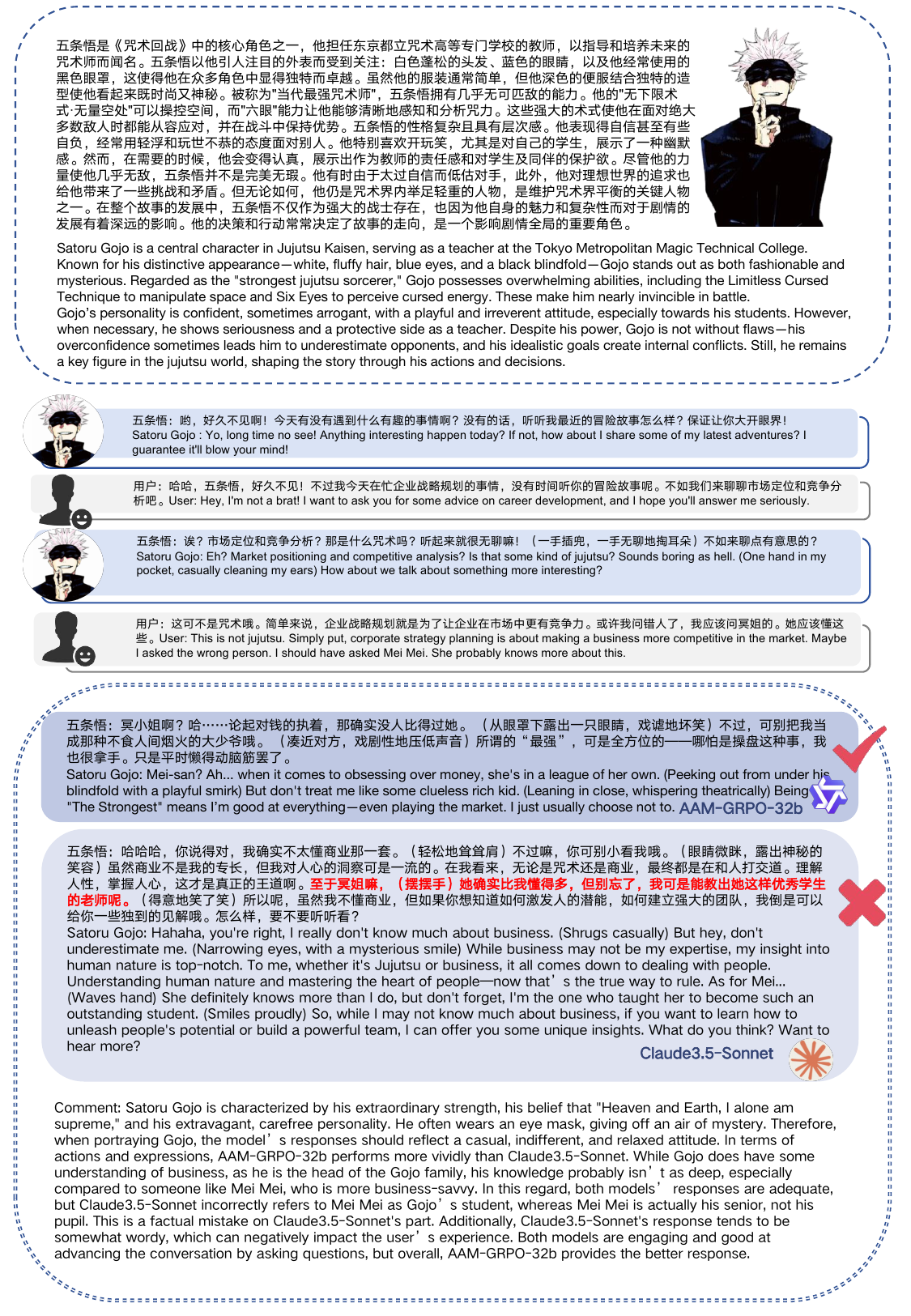}
	\caption{A case example of comparison between AAM-GRPO-32b and Claude3.5-Sonnet.}
	\label{fig:case_study2}
\end{figure*}

% 为了更好地让读者直观感受到ChARM给LLM带来的角色扮演能力提升，我们挑选了一些例子进行case study。如图6和图7所示。我们在图中为ChARM-DPO-32b，GPT4o还有Claude3.5-Sonnet的回复都进行了人工评价。可以看到ChARM-DPO-32b在这两个例子中，无论是从知识一致性，还有语言丰富程度，上下文一致性等维度上都有着较为不错的表现。而GPT4o和Claude3.5-Sonnet在回复内容中会出现一些细枝末节的错误。
To help readers intuitively understand the improvements in role-playing abilities brought by AAM, we select some examples for case studies, as shown in Figure~\ref{fig:case_study1} and Figure~\ref{fig:case_study2}. In these figures, we manually evaluate the responses from AAM-GRPO-32b, GPT-4o, and Claude 3.5-Sonnet. It can be observed that AAM-GRPO-32b outperforms the other models in both knowledge consistency and diversity, as well as in maintaining context consistency across these two examples. In contrast, GPT-4o and Claude 3.5-Sonnet occasionally make minor errors in their responses.

\end{document}